\PassOptionsToPackage{dvipsnames,table}{xcolor}
\documentclass[sigconf=true, nonacm=true, review=false, anonymous = false]{acmart}
\usepackage{booktabs} % For formal tables
\usepackage{graphicx}

\usepackage{multirow}
\usepackage{amsmath}
\usepackage{xspace}
\usepackage{xcolor}
\usepackage{dsfont}
\usepackage{rotating}
\usepackage{makecell}

\usepackage{comment}

\usepackage{amsthm}
\usepackage{subcaption}
\renewcommand{\epsilon}{\varepsilon}

\usepackage{mathtools}

\usepackage{tikz}
\usetikzlibrary{decorations.pathreplacing,calc,tikzmark}
\usepackage{pgfplots}
\pgfplotsset{compat=newest}
\pgfplotscreateplotcyclelist{tikzcycle}{%
thick,blue,mark=square*\\ % q = 1/n^2
thick,red,mark=*\\ % q = ln(n)/n^2
thick,green!70!black,mark=triangle*\\ % q = 1/n
thick,orange,mark=diamond*\\ % q = 1
thick,magenta,mark=x\\ % q = 1/n^2
thick,red,dashed,mark=o\\ % q = ln(n)/n^2
thick,green!70!black,dashed,mark=triangle\\ % q = 1/n
thick,orange,dashed,mark=diamond\\ % q = 1
}
\pgfplotscreateplotcyclelist{tikzcycle2}{%
thick,red,mark=*\\ % q = ln(n)/n^2
thick,green!70!black,mark=triangle*\\ % q = 1/n
}

\newcommand{\R}{\mathbb{R}}
\newcommand{\N}{\mathds{N}}

\newcommand{\calK}{\mathcal{K}}

\newcommand{\pleave}{p_{\text{leave}}}

\newcommand{\rlsk}{\textsc{RLS}$_k$\xspace}
\newcommand{\OneMax}{\textsc{OneMax}\xspace}
\newcommand{\LeadingOnes}{\textsc{LeadingOnes}\xspace}
\newcommand{\OM}{\textsc{OM}\xspace}
\newcommand{\LO}{\textsc{LO}\xspace}
\newcommand{\loom}{$(\LO, \OM)$\xspace}

\newcommand{\calS}{\mathcal{S}}
\newcommand{\calSone}{\mathcal{S}^{\LO}}
\newcommand{\calStwo}{\mathcal{S}^{(\LO, \OM)}}
\newcommand{\calSn}{\mathcal{S}^{x}}

\usepackage{multirow}
\usepackage[linesnumbered,ruled]{algorithm2e}

\DeclareMathOperator{\Bin}{Bin}
\DeclareMathOperator{\HG}{HG}

\begin{document}
\title{Enhancing Parameter Control Policies with State Information}

\author{Gianluca Covini}
\orcid{0009-0001-2803-8209}
\affiliation{
  \institution{University of Pavia}
  \city{Pavia}
  \country{Italy}}
% \email{gianluca.covini01@universitadipavia.it}

\author{Denis Antipov}
\orcid{0000-0001-7906-096X}
\affiliation{
  \institution{Sorbonne Universit\'e, CNRS, LIP6}
  \city{Paris}
  \country{France}}
% \email{denis.antipov@lip6.fr}

\author{Carola Doerr}
\orcid{0000-0002-4981-3227}
\affiliation{
  \institution{Sorbonne Universit\'e, CNRS, LIP6}
  \city{Paris}
  \country{France}}
% \email{carola.doerr@lip6.fr}

% \renewcommand{\shortauthors}{}

\begin{abstract} 
Parameter control and dynamic algorithm configuration study how to dynamically choose suitable configurations of a parametrized algorithm during the optimization process. Despite being an intensively researched topic in evolutionary computation, optimal control policies are known only for very few cases, limiting the development of automated approaches to achieve them.

With this work we propose four new benchmarks for which we derive optimal or close-to-optimal control policies. More precisely, we consider the optimization of the \LeadingOnes function via RLS$_{k}$, a local search algorithm allowing for a dynamic choice of the mutation strength $k$. The benchmarks differ in which information the algorithm can exploit to set its parameters and to select offspring. In existing running time results, the exploitable information is typically limited to the quality of the current-best solution. In this work, we consider how additional information about the current state of the algorithm can help to make better choices of parameters, and how these choices affect the performance. Namely, we allow the algorithm to use information about the current \OneMax value, and we find that it allows much better parameter choices, especially in marginal states. Although those states are rarely visited by the algorithm, such policies yield a notable speed-up in terms of expected runtime. This makes the proposed benchmarks a challenging, but promising testing ground for analysis of parameter control methods in rich state spaces and of their ability to find optimal policies by catching the performance improvements yielded by correct parameter choices.   
\end{abstract}

\begin{CCSXML}
<ccs2012>
         <concept>
      <concept_id>10003752.10003809.10003716.10011138.10011803</concept_id>
      <concept_desc>Theory of computation~Bio-inspired optimization</concept_desc>
      <concept_significance>500</concept_significance>
      </concept>
 </ccs2012>
\end{CCSXML}

% \ccsdesc[500]{Computing methodologies~Continuous space search}
\ccsdesc[500]{Theory of computation~Bio-inspired optimization}
% \ccsdesc[500]{Applied computing~Multi-criterion optimization and decision-making}

\keywords{Parameter control, black-box optimization, evolutionary computation, local search, algorithm configuration}

\maketitle

\sloppy
\section{Introduction}
\label{sec:intro}

The efficiency of optimization algorithms heavily depends on the choice of their parameters. While evolutionary algorithms and other black-box optimization techniques have been widely used across different problem domains, the question of how to optimally set algorithm parameters remains an ongoing challenge. A common approach is \textit{parameter control}, where algorithm parameters are adjusted dynamically during the optimization process rather than being fixed beforehand~\cite{DoerrD18chapter,KarafotiasHE15,DoerrDL21, Doerr19domi}. 

Recently, a novel framework called \emph{Dynamic Algorithm Configuration (DAC)} has gained increasing attention as an approach to enhance the performance of optimization algorithms by adjusting their parameters in response to changing problem conditions~\cite{Adriansen2022DACjournal, SharmaKLK19ManuelDACRL, DeyaoGECCO23}. The main difference of DAC from parameter control is that it allows for an explicit training phase to identify good control policies, i.e., a mapping from the current state to suitable parameter values, while in parameter control parameter values have to be chosen on-the-fly, i.e. during the optimization process. Reinforcement Learning (RL) has emerged as a promising technique for DAC, as it allows an agent to learn adaptive parameter control strategies from experience. An RL-based parameter approach has been applied in~\cite{BiedenkappDKHD22} to the radius choice of Randomized Local Search (RLS) algorithm on the problem \LeadingOnes showing promising results. However, one of the main challenges that emerged with using RL for parameter control and DAC is the difficulty of generalizing the learning behavior to high-dimensional settings where the current state of the optimization process is captured by more than one numerical value (in many parameter control studies, the current state is exclusively described by the quality of the current-best solution~\cite{BottcherDN10,DoerrDY20,Doerr19domi}). In particular, we note that static or longitudinal information about the evaluated solutions and their fitness values are currently not considered, or only in highly aggregated form such as through success-based rules~\cite{DoerrDY16PPSN,DoerrD18ga,DoerrDL21,FajardoS22,FajardoS24ai,FajardoS24algo}. 

Development and understanding of parameter control and DAC methods, especially the ones that are based on RL, heavily relies on benchmarks, where they can be analyzed and where their policy of choosing parameters can be evaluated against a known optimal policy. 
With the desire to define benchmarks allowing to study more complex parameter control policies (whether learned online or through a training process), we investigate  settings in which richer state information is available. For this we need new benchmarks with higher-dimensional state spaces that satisfy the following requirements:
\begin{enumerate}
    \item[(1)] we need to be able to compute the optimal policy of choosing parameters for each state of the state space;
    \item[(2)] the wrong parameters choices should worsen the performance of the algorithm;
    \item[(3)] preferably the optimal parameters should vary depending on the stage of optimization so that dynamic parameter control had clear advantage over static parameters tuning.    
\end{enumerate} 
Recent studies have suggested that additional state information can sometimes enhance policy learning, but its exact impact on optimization efficiency remains unclear. As a negative example, \citet{Buzdalov2015COHOLO} considered optimization of \LeadingOnes with auxiliary objective \OneMax which could be chosen to be optimized by an RL algorithm, and they showed that smaller state spaces allow to find the optimum faster.

In this work we propose several benchmarks that are based on RLS and \LeadingOnes and we explore whether incorporating state representations beyond \LeadingOnes fitness can improve dynamic parameter policies. Specifically, we study the optimization of \LeadingOnes with three different state descriptors: \LeadingOnes-values only, \LeadingOnes and \OneMax values, and the current-best solution $x$. We consider different combinations of using this state information in two parts of the algorithm: in the choice of the distance to sample an offspring (i.e., the choice of the mutation strength $k$) and in the selection of the next-generation parent, when a current parent and its offspring compete with each other.

We compute exact optimal or close-to-optimal mutation rates under different state space representations and we show that the additional information improves the performance compared to the using only a limited information about the current \LeadingOnes value. This improvement is most pronounced when we use these clues in the selection step, giving an asymptotical speed-up, but we also see a notable improvement when we only use additional information for choosing mutation strength, especially when we use strict selection in the RLS (that is, when we only accept offspring if its \LeadingOnes value is strictly better than its parent's).
Our findings provide valuable insights into the conditions under which additional state information is beneficial, refining DAC strategies and contributing to the design of more effective adaptive algorithms.

The remainder of the paper is organized as follows. Section~\ref{sec:preliminaries} introduces the problem formulation. Section~\ref{sec:loom} describes the approaches developed to compute the policy of choosing mutation strength when we use the state information in the selection step in cases when the state information includes \OneMax values or the full information about the current bit string $x$. Section~\ref{sec:lo} presents policies when we do not use additional information in selection, but we use the current \OneMax value when selecting the search radius of the RLS. Section~\ref{sec:portfolio} presents results when limiting the portfolio of possible parameter values, Section~\ref{sec:summary} concludes with a discussion of our findings and directions for future research.

\textbf{Reproducibility.} The code that was used for numerical computation and for experiments described in this paper can be found on GitHub~\cite{github}.

\section{Preliminaries}\label{sec:preliminaries}

In this section we introduce the notation that we use in this paper, and the three main components of the proposed benchmarks: the algorithm, the problems, and the state spaces for parameter control. We then explain our goals that we stated in the introduction in the context of formally defined benchmarks.

In this paper, for two integer numbers $a$ and $b$ ($b \ge a$) by $[a..b]$ we denote an integer interval, that is, all integer numbers that are at least $a$ and at most $b$. For any $m \in \N$ by $1^m$ and $0^m$ we denote a bit string of $m$ ones and $m$ zeros respectively. For $m = 0$ both denote an empty bit string.

\subsection{The Random Local Search}

As the main subject for testing parameter control methods, we consider the random local search with variable search radius $k$, which we denote by \rlsk. This is an algorithm for pseudo-Boolean optimization, that is, in the space $\{0, 1\}^n$ of bit strings of length $n$. It stores one individual $x$ that is initialized with a random bit string, and in each iteration it creates an offspring $y$ by first choosing radius of search $k$ and then flipping bits of $x$ in exactly $k$ positions that are chosen uniformly at random. If the value of the optimized function in $y$ is not worse than in $x$, then $y$ replaces $x$, and otherwise $x$ stays the same. These iterations are repeated until some stopping criterion is met. 
The pseudocode of the \rlsk is shown in Algorithm~\ref{alg:rlsk}.

We do not specify the stopping criterion and assume that the algorithm never stops before finding a global optimum of the optimized function. We measure the performance of the algorithm as the number of fitness evaluations that it makes before it evaluates a global optimum for the first time, and we call it the algorithm's \emph{runtime}. This is a random variable (due to the random choices during the algorithm's run), hence our main interest is in the \emph{expected runtime}: the smaller it is, the better is the performance of the algorithm.

The choice of the search radius in each iteration is made based on the state of the algorithm in the beginning of that iteration. The state of the algorithm at any time is defined by the current individual $x$, that is, we can consider the state as some function $S: \{0, 1\}^n \mapsto \calS$, where $\calS$ is the state space. We discuss the state spaces that we consider in Section~\ref{sec:states}. 

Sometimes the algorithm has a limited set of search radii it can use. We call this set a \emph{portfolio} and denote it by $\calK$. By default, we assume that $\calK = [1..n]$, that is, it includes all possible options, but in Section~\ref{sec:portfolio} we study settings with limited portfolios.

In Sections~\ref{sec:loom-sn} and~\ref{sec:lo-strict} we also consider a modified version of the \rlsk which accepts new offspring only if it is \emph{strictly better} than the parent. That is, in line~\ref{line:comparison} of Algorithm~\ref{alg:rlsk} we use a strict inequality sign. Although it does not usually make sense for the real-world optimization (since it significantly shrinks the ability of the algorithm to explore the search space), our goal is to make a good testing ground for parameter control mechanisms, and we observe that this selection yields a more notable benefit from richer state space. 

\begin{algorithm}[t!]%
	\caption{The \rlsk maximizing $f:\{0,1\}^n \rightarrow \R$.}
	\label{alg:rlsk}
		Sample $x \in \{0,1\}^{n}$ uniformly at random\;
		
		\While{not stopped}{
			$s \gets S(x)$ \tcp*{Computing the current state}
			$k \gets k(s)$ \tcp*{Choosing $k$ based on $s$}
            $y \gets$ copy of $x$\;
            Flip $k$ bits in $y$ in positions chosen u.a.r.\;
			\If{$f(y) \ge f(x)$}{\label{line:comparison}
				$x \gets y$\;
			} %if
		} %while 
\end{algorithm}

\subsection{Fitness Functions}
\label{sec:fitness}

We consider fitness functions that are based on classic benchmarks \LeadingOnes and \OneMax. \LeadingOnes (\LO for brevity) was first proposed in~\cite{Rudolph97}. It returns the size of the longest prefix consisting only of one-bits in its arguments. More formally, for any bit string $x \in \{0, 1\}^n$ we have
\begin{align*}
    \LeadingOnes(x) = \LO(x) = \sum_{i = 1}^n \prod_{j = 1}^i x_i.
\end{align*}

\OneMax (\OM for brevity) is the second benchmark that we use in this work, and it simply returns the number of one-bits in its argument, that is,
\begin{align*}
    \OneMax(x) = \OM(x) = \sum_{i = 1}^n x_i.
\end{align*}

Another fitness function which we denote by \loom combines both \OneMax and \LeadingOnes and for any bit sting $x$ it returns a tuple $(\LO(x), \OM(x))$. When we compare two individuals based on this fitness function (in line~\ref{line:comparison} of Algorithm~\ref{alg:rlsk}), these tuples are compared in lexicographical order, that is, we first compare the values of \LeadingOnes, and if they are equal, then we compare \OneMax values. This function can also be represented as a scalar function that returns $(n + 1)\LO(x) + \OM(x)$, but we prefer to consider it as a tuple, since it highlights the idea of having \LeadingOnes as the target function, and using \OneMax just as an auxiliary objective which helps the optimization when there is no signal from the main objective.

We note that the \rlsk can optimize \loom in $O(n\log n)$ time, which is asymptotically faster than the best possible runtime of mutation-only algorithms on \LeadingOnes, which is $\Omega(n^2)$~\cite{LehreW12}. This upper bound for \loom follows from a simple observation that the standard RLS that always flips one bit behaves on \loom similar to \OneMax: the fitness increases when it flips zero to one, and the fitness decreases otherwise. Hence it essentially optimizes \OneMax and finds the optimum in $\Theta(n\log(n))$ iterations in expectation. The performance of the \rlsk with optimal policy for choosing $k$ cannot be worse than this, hence it solves \loom in $O(n\log(n))$ time.

We also note that it has been shown in~\cite{Buzdalov2015COHOLO} that \OneMax can help to optimize \LeadingOnes when used as a secondary objective. However this is quite an artificial combination of functions. In real-world problems it is also necessary to detect helping objectives among auxiliary objectives that might be available, which is often solved with RL algorithms~\cite{BuzdalovaB12,BuzdalovB13}. For us, however, this artificiality is not important, since the goal of our benchmarks is to make a testing ground for dynamic methods of parameter configuration, not for optimization algorithms.

\subsection{State Spaces and Transition Probabilities}
\label{sec:states}

The \rlsk chooses the search radius $k$ depending on the current state of the algorithm that is defined by the current individual $x$. That is, the state space can be seen as a partition of the search space into disjoint sets of bit strings, and each of these sets represents a state of the algorithm. For any state $s$ we say that the \rlsk is in $s$, if the current individual $x$ belongs to $s$. We consider three different state spaces.

\textbf{The first state space} $\calSone$ is based solely on the \LeadingOnes values of the points in the search space. It consists of $n + 1$ states $s_0, \dots, s_n$, and for all $i \in [0..n]$ state $s_i$ consists of all bit strings $x$ with $\LO(x) = i$. For this state space, we denote by $p_i^j(k)$ (where $i, j \in [0..n]$) the probability that the \rlsk creates an offspring from $s_j$ conditional on being in $s_i$ and flipping exactly $k$ bits. Note that depending on the optimized function (just \LeadingOnes or \loom) the \rlsk might either accept this offspring (and go to a new state) or not accept it (and stay in the same state).

\textbf{The second state space} $\calStwo$ is based on both \LeadingOnes and \OneMax. Its consists of states $(s_{i, j})_{i, j \in [0..n]}$, where state $s_{i, j}$ consists of all bit strings $x$ with $\LO(x) = i$ and $\OM(x) = j$. Since the \OneMax value of any bit string cannot be less than its \LeadingOnes value, all states $s_{i, j}$ with $i > j$ are empty. Also, since $\OM(x) = n$ only for the all-ones bit string, states $s_{i, n}$ with $i < n$ are also empty. In the rest of this paper we pretend that these empty states do not exist. For this state space, we denote by $p_{i, j}^{\ell, m}(k)$ (where $i, \ell \in [0..n]$, $j \in [i..n]$ and $m \in [\ell..n]$) the probability that the \rlsk creates an offspring from state $s_{\ell, m}$ conditional on being in state $s_{i, j}$ and flipping exactly $k$ bits (again, it does not guarantee that the \rlsk goes to state $s_{\ell, m}$ due to selection).

To compute $p_{i, j}^{\ell, m}(k)$, we note that when the algorithm is in state $s_{i, j}$, the current individual $x$ starts with $1^i0$ and then has exactly $j - i$ one-bits in positions $[i + 2..n]$. The positions of one-bits in the suffix are distributed uniformly at random, which can be shown by the arguments similar to~\cite{Rudolph97} (roughly speaking, any combination of positions is equiprobable, since any trajectory leading to $s_{i, j}$ does not give us information about suffix, except for the number of one-bits in it). Keeping it in mind, we distinguish two cases of computing $p_{i, j}^{\ell, m}(k)$.

First, when $\ell = i$, then $p_{i, j}^{\ell, m}(k)$ is the probability that we flip $k$ bits so that (i) we do not flip bits in positions $[1..i+1]$ (the \LO value stays the same) and (ii) in positions $[i + 2..n]$ we flip $k_0$ zero-bits and $k_1$ one-bits so that the change of \OM value $k_0 - k_1$ is exactly $m - j$, which together with $k_0 + k_1 = k$ (which follows from (i)) implies that $k_0 = \frac{m - j + k}{2}$. The probability of (i) is $\binom{n - i - 1}{k} / \binom{n}{k}$, and conditional on (i), $k_0$ follows the hyper-geometric distribution $\HG(n - i - 1, n - j - 1, k)$, hence
\begin{align*}
    p_{i, j}^{i, m}(k) = \frac{\binom{n - i - 1}{k}}{\binom{n}{k}} \cdot \frac{\binom{n - j - 1}{\frac{m - j + k}{2}} \binom{j - i}{k - \frac{m - j + k}{2}}}{\binom{n - i - 1}{k}} = \frac{\binom{n - j - 1}{\frac{m - j + k}{2}} \binom{j - i}{\frac{k - m + j}{2}}}{\binom{n}{k}}. 
\end{align*}
When $\ell \ge i$, then $p_{i, j}^{\ell, m}(k)$ is the probability that we flip $k$ bits so that (i) we do not flip bits in positions $[1..i]$ and flip a zero-bit in position $i + 1$ (the \LO value increases), (ii) we flip $k_0 = \frac{m - j + k - 1}{2}$ zero-bits in positions $[i + 2..n]$ and (iii) after it we get one-bits in positions $[i + 2..\ell]$ and a zero-bit in position $\ell + 1$. The probabilities of (i) and (ii) can be computed as in the previous case for $\ell = i$, with the only change that we must flip the zero-bit in position $i + 1$, and hence we flip $k - 1$ bits in the suffix $[i + 2..n]$. For estimating the probability of (iii) we note that conditional on (i) and (ii) we have $m - i - 1$ one-bits in positions $[i + 2..n]$, and the positions of these one-bits are distributed uniformly at random, as we have discussed. Therefore, the probability that the new \LO value is $\ell$ is the probability that the bits in positions $[i + 2..\ell]$ are one-bits and the bit in position $\ell + 1$ is a zero-bit, that is $\binom{n - \ell - 1}{m - \ell} / \binom{n - i - 1}{m - i - 1}$. Hence, we have
\begin{align*}
    p_{i, j}^{i, m}(k) &= \frac{\binom{n - i - 1}{k - 1}}{\binom{n}{k}} \cdot \frac{\binom{n - j - 1}{\frac{m - j + k - 1}{2}} \binom{j - i}{(k - 1) - \frac{m - j + k - 1}{2}}}{\binom{n - i - 1}{k - 1}} \cdot  \frac{\binom{n - \ell - 1}{m - \ell}}{\binom{n - i - 1}{m - i - 1}} \\
    &= \frac{\binom{n - j - 1}{\frac{m - j + k - 1}{2}} \binom{j - i}{\frac{k - 1 - m + j}{2}}}{\binom{n}{k}} \cdot  \frac{\binom{n - \ell - 1}{m - \ell}}{\binom{n - i - 1}{m - i - 1}}.
\end{align*}

\textbf{The third state space} $\calSn$ is the largest one: each point in the search space corresponds to its unique state. That is, for each $x \in \{0, 1\}^n$ there exists a state $s_x$, and this state consists of the single search point $x$. When we consider the \rlsk in this state space, it essentially means that it can choose $k$ based on the bit string $x$, but not only on its \LO and \OM values, which should give the best opportunities for a proper selection of the parameters. For any pair of bit strings $x$ and $y$ by $p_{x \to y}(k)$ we denote the probability to create $y$ via a $k$-bits flip from $x$. Since the $k$ bits to flip are chosen uniformly at random, this probability is
\begin{align*}
    p_{x \to y}(k) &= \begin{cases}
        \binom{n}{k}^{-1}, &\text{ if } H(x, y) = k \\
        0, &\text{ if } H(x, y) \ne k.
    \end{cases}
\end{align*}

\subsection{Problem Statement}
\label{sec:problem-statement}

The main goal of this paper is to design new benchmarks with rich state spaces for analysis of parameter control methods. For this we also want to find the optimal policies $\pi: \calS \mapsto \calK$ of choosing parameters from portfolio depending on the current state to use as a reference for tested DAC and parameter control methods. We also aim to understand how state information that goes beyond only the current fitness knowledge can help to choose better parameters. To achieve these goals, we propose four benchmark settings for choosing search radius $k$ in the \rlsk. These settings are listed in Table~\ref{tbl:list}.

\begin{table}[h]
    \centering
    \begin{tabular}{ccc}
    \toprule
        Fitness function & Offspring selection & State information   \\\midrule
        \multirow{2}{*}{\loom}  &
        Non-strict &
        \LO(x), \OM(x)  \\\cmidrule(lr){2-3}
        
        & Strict &
        bit string $x$\\\midrule

        \multirow{2}{*}{\LeadingOnes}  &
        Non-strict &
        \LO(x), \OM(x) \\\cmidrule(lr){2-3}
        
        & Strict &
        \LO(x), \OM(x) \\
        
    \bottomrule
    \end{tabular}
    \caption{The four proposed benchmark settings.}
    \label{tbl:list}
\end{table}

Benchmarks that include optimization on \loom also help us to study the effect of using additional information about the problem for selection, while in settings with \LeadingOnes we restrict the usage of the additional information and only use it to choose $k$.

In the next two sections we show how to compute the optimal policies for these settings and discuss how it can be used for analysis of DAC methods.

\section{Optimization of \loom}
\label{sec:loom}

In this section, we study the proposed benchmarks where the optimized function is \loom and the state spaces are $\calStwo$ (when we can choose $k$ based on the current fitness) and $\calSn$ (when we can choose $k$ based on the current bit string $x$). We describe the ways to compute the optimal policies for choosing $k$ depending on the current state precisely and show the runtime which these policies yield.
The main insight from this section is that the additional information used by the optimizer both for parameter selection and for offspring selection can be very beneficial for optimization, even reducing the runtime asymptotically. 

Before we start, it is important to note that when the \rlsk optimizes \loom, it never decreases the \LO value of its current individual $x$, and it also never decreases \OM value when keeping the same \LO value. This observation allows us to define the order of states in $\calStwo$, where we say that state $s_{i, j} > s_{\ell, m}$ if either $i > \ell$ or if $i = \ell$ and $m > j$. With this order, the \rlsk can only go from $s_{i, j}$ to $s_{\ell, m}$, if and only if $s_{i, j} \le s_{\ell, m}$.

\subsection{State Space of \LO and \OM Values}
\label{sec:loom-s2}

In this section we use $\calStwo$ state space for choosing the search radius $k$. We first compute the optimal policy for this setting, then compare it with some other sub-optimal policies and then also compare the runtimes that these policies yield.

\textbf{Computation of the optimal policy.} Before we show the optimal policy of choosing $k$, we describe how we can compute the expected runtimes when we know the policy of choosing $k$ for all states in $\calStwo$.
Assume that we have a set $K = (k_{i, j})_{i \in [0..n - 1], j \in[i..n-1]}$, and a policy which chooses $k = k_{i, j}$ when the \rlsk in state $s_{i, j}$ for all $i$ and $j$ (recall that states with $j < i$ and states with $i < n \land j = n$ do not exist, and the state with $i = j = n$ is the optimal state, hence it does not matter how many bits we flip in it). Let also, for all $i \in [0..n - 1]$ and $j \in[i..n-1]$ $T_{i, j}(K)$, be the number of iterations it takes the \rlsk to reach the all-ones bit string when starting in state $s_{i, j}$ and flipping bits according to the policy defined by $K$. We also define $T_{n, n}(K) = 0$ for any $K$. Then, we can use the order of the states of $\calStwo$ to compute the expected runtimes for each space. 

We start by computing the runtime for state $s_{n - 1, n - 1}$. If the algorithm is in this state, then the current bit string $x$ must be $1^{n - 1}0$ (that is, $n - 1$ ones followed by a single zero). Hence, if $k_{n - 1, n - 1} = 1$, the algorithm generates the optimum with probability $\frac{1}{n}$ by flipping the only zero-bit, and the expected runtime is $E[T_{n - 1, n - 1}(K)] = n$. Otherwise, if $k_{n - 1, n - 1} \ne 1$, then the algorithm cannot leave the state, since it can only decrease the \LO value with two- or more-bits flips. Hence, $E[T_{n - 1, n - 1}(K)] = +\infty$. 

After finding $E[T_{n - 1, n - 1}(K)]$, we can compute the expected runtimes for all other states, iterating through them in descending order. Consider computation of $E[T_{i, j}(K)]$ assuming that we have already computed the expected runtimes for states that are larger than $s_{i, j}$ (according to the order defined in the beginning of this section). Define $\pleave$ as the probability of leaving state $s_{i, j}$ in one iteration when flipping $k_{i,j}$ bits. Then if $\pleave = 0$, we have $E[T_{i, j}(K)] = +\infty$, as we never leave state $s_{i, j}$. Otherwise, if $\pleave > 0$, then the following equation on the expected runtimes and its transformations allow us to compute $E[T_{i, j}(K)]$.

\begin{align}\label{eq:ET-loom-s2}
    \begin{split}
        E[T_{i, j}(K)] &= 1 + \sum_{(\ell, m): s_{\ell, m} > s_{i, j}} p_{i, j}^{\ell, m}(k_{i, j}) E[T_{\ell, m}(K)] \\
        &+ (1 - \pleave) E[T_{i, j}(K)]; \\
        \pleave E[T_{i, j}(K)] &= 1 + \sum_{(\ell, m): s_{\ell, m} > s_{i, j}} p_{i, j}^{\ell, m}(k_{i, j}) E[T_{\ell, m}(K)]; \\
        E[T_{i, j}(K)] &= \frac{1 + \sum_{(\ell, m): s_{\ell, m} > s_{i, j}} p_{i, j}^{\ell, m}(k_{i, j}) E[T_{\ell, m}(K)]}{\pleave}.
    \end{split}
\end{align}
The probability $\pleave$ of leaving the state is the sum of the transition probabilities from $s_{i, j}$ to all larger states, including the optimal state $s_{n,n}$, that is,
\begin{align*}
    \pleave = \sum_{(\ell, m): s_{\ell, m} > s_{i, j}} p_{i, j}^{\ell, m}(k_{i, j}).
\end{align*}

This approach also allows us to find the best values of $k_{i, j}$ deductively. Assume that we have already computed the optimal values of $k$ for all states larger than $s_{i, j}$. Then to find the best $k_{i, j}$ we need to compute eq.~\eqref{eq:ET-loom-s2} for all available values of $k_{i, j}$ and choose the one that minimizes $E[T_{i, j}(K)]$.

\begin{figure}
    \includegraphics[height=3.55cm]{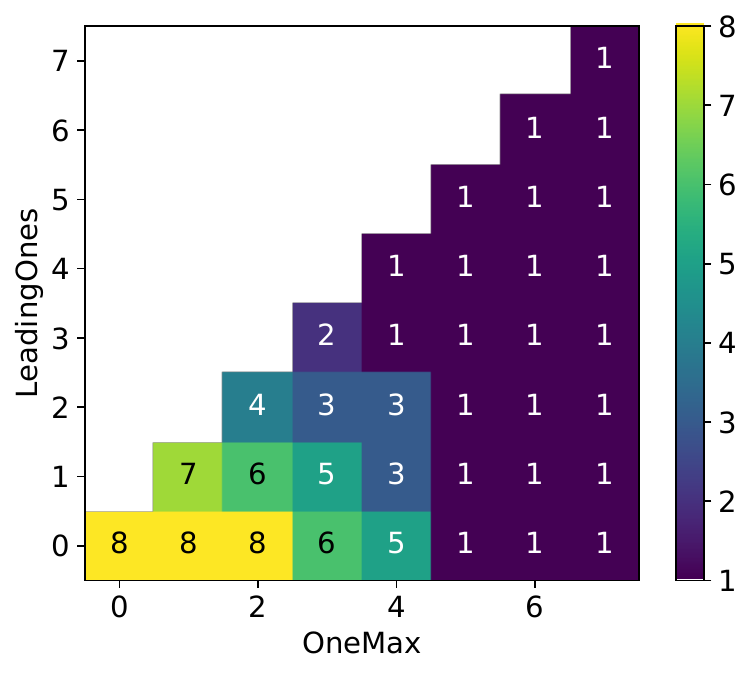}
    \includegraphics[height=3.5cm]{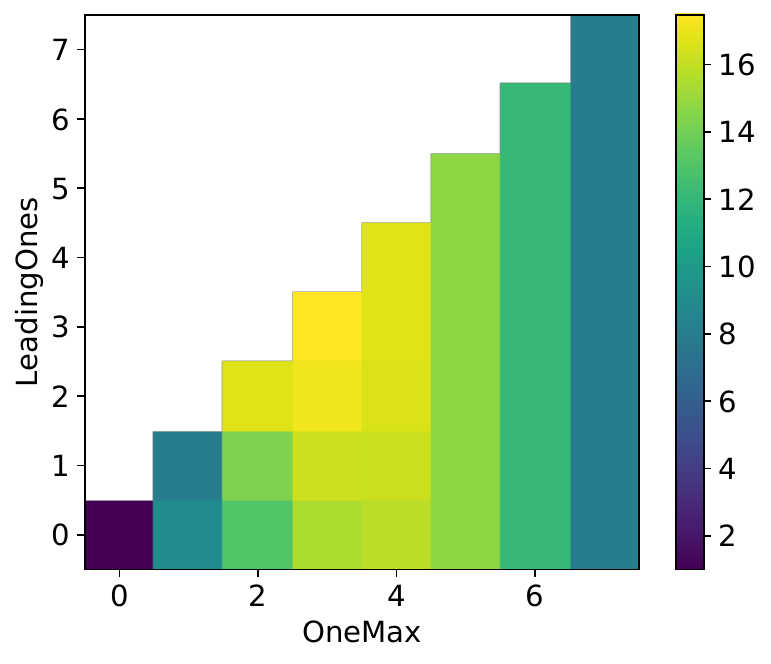}
    \caption{An example heatmap showing the optimal search radii (left) and the optimal runtimes (right) for the \rlsk optimizing \loom with $n = 8$ which can choose the search radius $k$ based on the \loom value (that is, based on its state from state space $\calStwo$).}
    \label{fig:example}
\end{figure}

\begin{figure*}
    \includegraphics[width=0.33\linewidth]{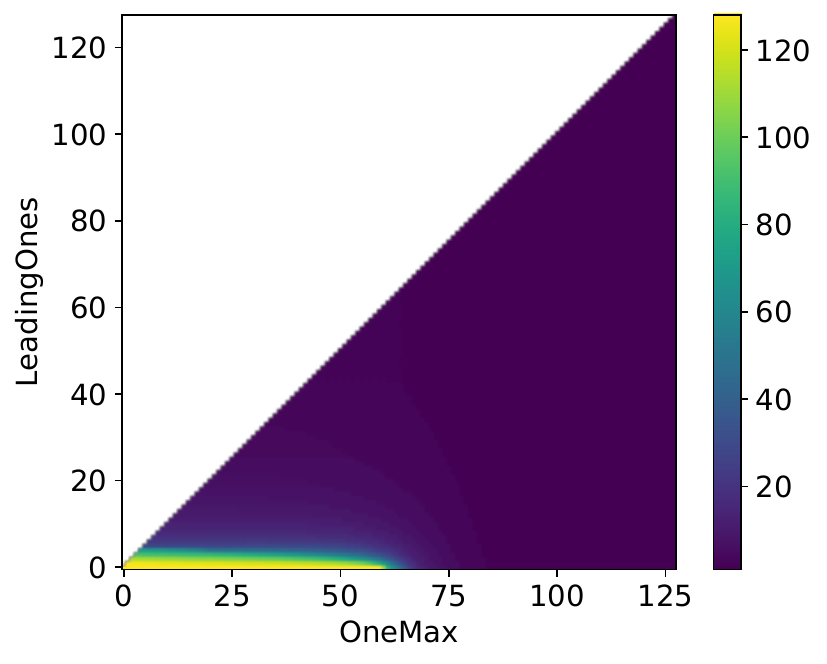}
    \includegraphics[width=0.33\linewidth]{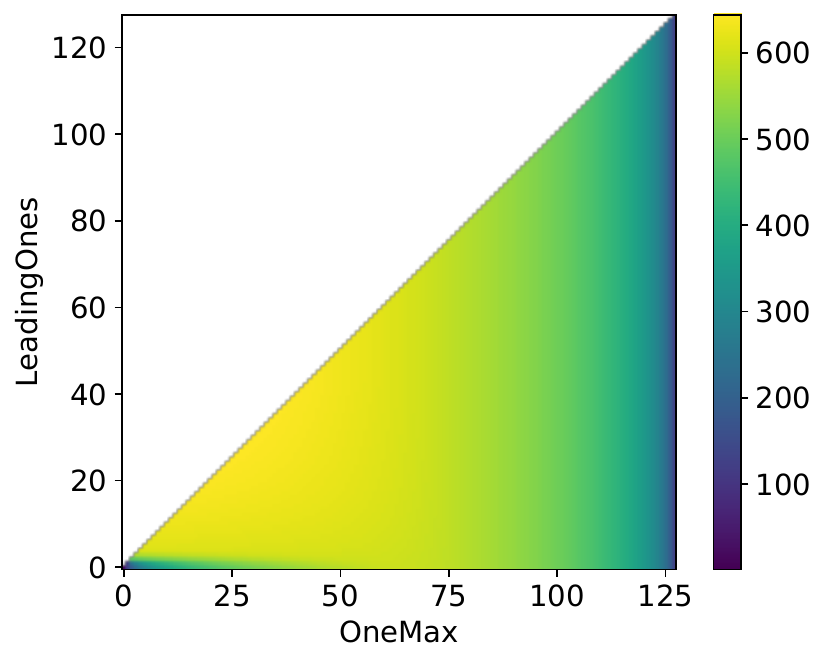}
    \includegraphics[width=0.33\linewidth]{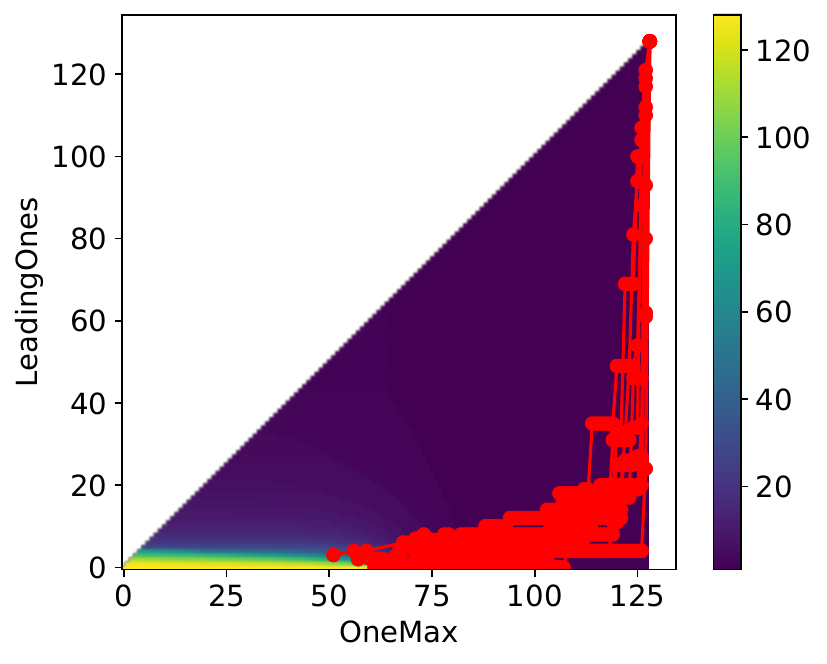}
    \caption{Heatmaps for the \rlsk on \loom using state space $\calStwo$: optimal search radii (left), expected runtimes (middle) and trajectories of 10 typical runs over the optimal rates heatmap (right). The problem size is $n = 128$.}
    \label{fig:loom-s2}
\end{figure*}

\begin{figure*}
    \includegraphics[width=0.33\linewidth]{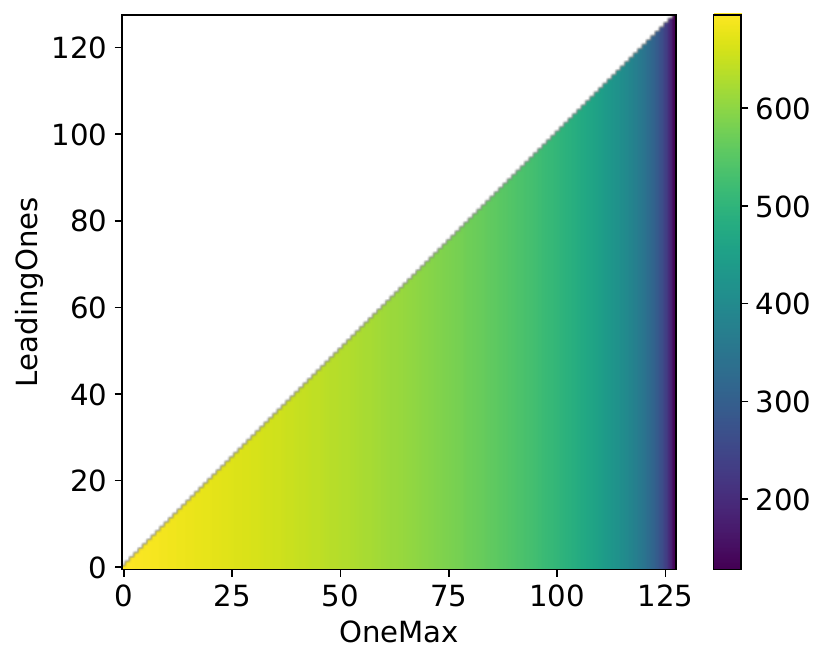}
    \includegraphics[width=0.33\linewidth]{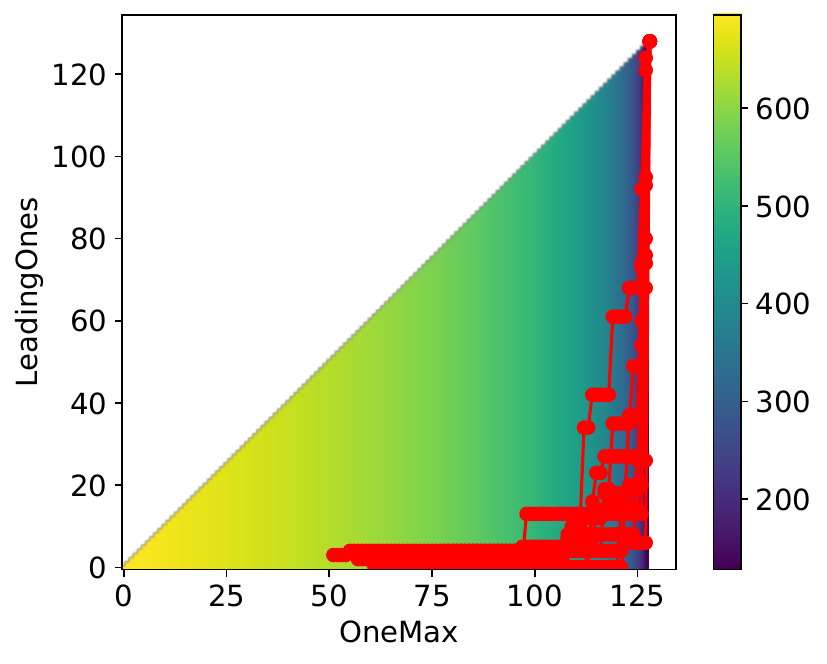}
    \caption{Heatmaps for the the RLS (always flipping one bit) on \loom: expected runtimes (left) and trajectories of 10 typical runs over the runtimes heatmap (right). The problem size is $n = 128$.}
    \label{fig:loom-s2-onebit}
\end{figure*}

\textbf{Comparison of different policies.} To illustrate the optimal policy $K$ that we computed and the runtimes that this policy yields, we use heatmaps. A small example is shown in Figure~\ref{fig:example}. In this figure, the left plot shows a heatmap of the optimal rates in the $\calStwo$ state space. Each state of the algorithm is defined by the current \LO value (Y-axis) and \OM value (X-axis). The numbers show the optimal number of bits to flip in each state, and it is also shown via colors: the darker squares mean lower search radii. We will omit precise numbers in larger figures. The right plot shows the expected runtimes for each state, that is, how much time the \rlsk would need to find the all-ones bit string if it starts in each state. As one can see in this figure, the easiest state is $(0, 0)$ (since we exclude the state $(n, n)$ with the optimum), from which the algorithm can find the optimum in one iteration by flipping all bits, while it takes the algorithm the most time to find the optimum, when it starts with $\LO = \OM = 3$ : the brighter regions correspond to larger runtimes.

Figure~\ref{fig:loom-s2} shows similar heatmaps for larger problem size $n = 128$. We see in this figure that in most states when $\OM > \frac{n}{2}$ it is optimal to flip only one bit. This gives the algorithms the best ability to optimize \OneMax without decreasing its value (e.g., when it improves \LO value). This is different for states with low \LO value and with $\OM < 0.4n$: there it is better to flip lots of bits, since hopefully it can bring us to a much better state in one iteration. The expected runtimes also reflect that the algorithms tend to optimize \OM, even though it is the second objective: the gradient of color goes almost parallel to X axis. The trajectories of ten runs also support the observation that the algorithm tries to optimize \OM first. We see that at start the trajectories go right and slightly up, but then, when the tail of the algorithm is full of one-bits, we see large improvements in \LO values as the trajectories go steeply up. We also see that the hardest state to start the run is when $\LO \approx n/3$ and $\OM \approx \LO$ (that is, we have many zero-bits in the tail).

\begin{figure*}
    \includegraphics[width=0.33\linewidth]{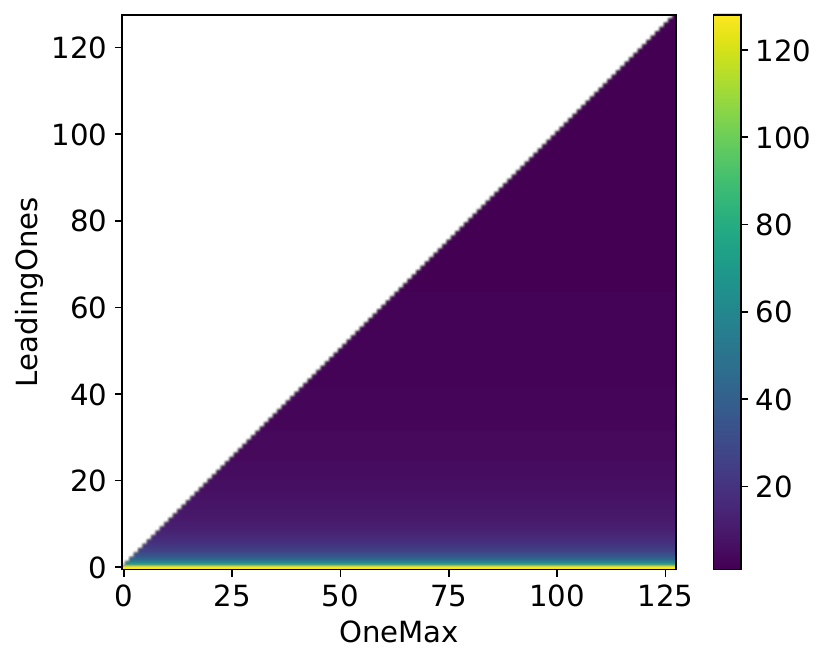}
    \includegraphics[width=0.33\linewidth]{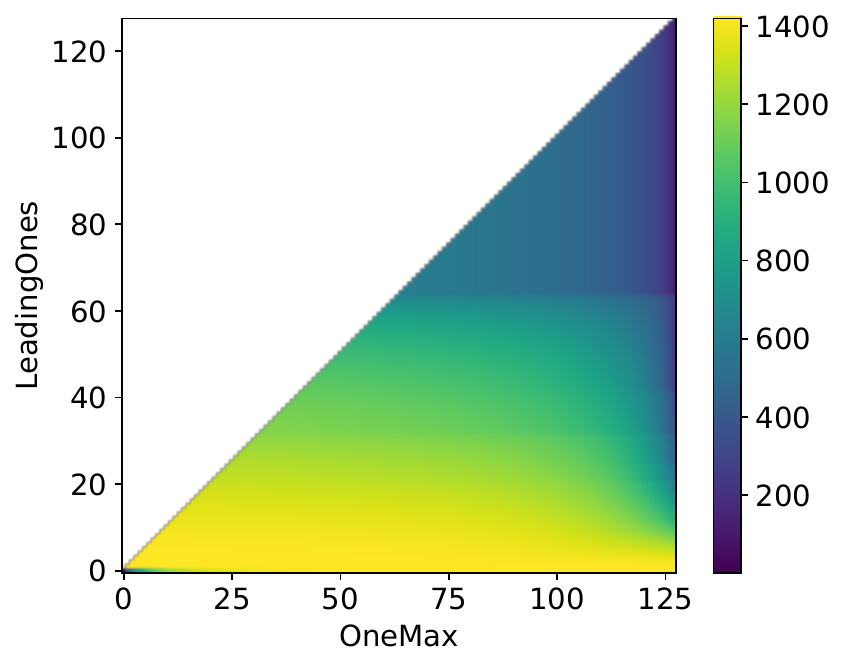}
    \includegraphics[width=0.33\linewidth]{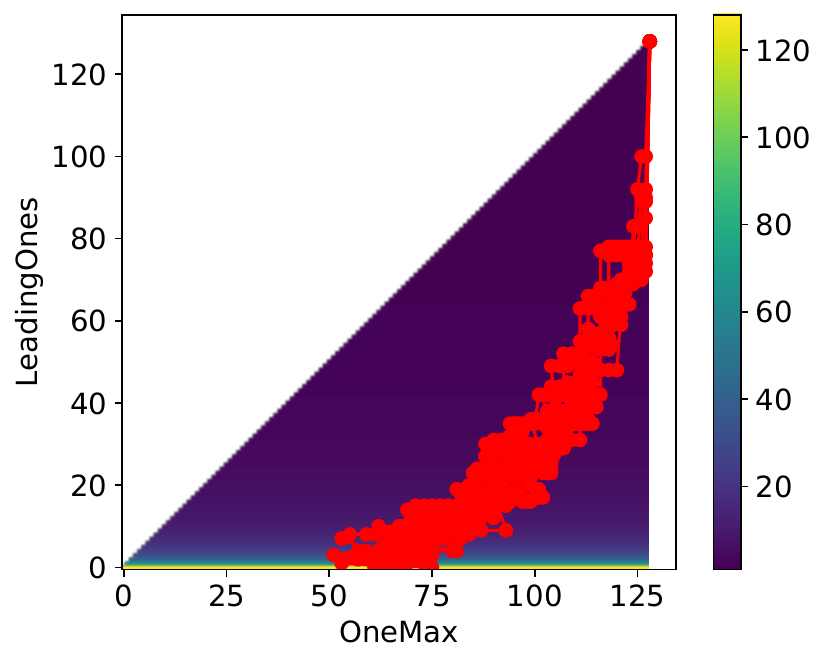}
    \caption{Heatmaps for the \rlsk on \loom that chooses radii optimal for \LeadingOnes optimization in state space $\calSone$: optimal search radii (left), expected runtimes (middle) and trajectories of 10 typical runs over the optimal rates heatmap (right). The problem size is $n = 128$.}
    \label{fig:loom-s1-loopt}
\end{figure*}

In Figure~\ref{fig:loom-s2-onebit}, we present similar heatmaps for a policy when $k$ is always one for comparison. Similar to the runs with optimal policy $K$, the trajectories first increase \OM fast, and then do several large jumps up. Note that in this case the algorithm cannot decrease \OM: even if it increases \LO value, it does it by flipping one zero-bit, hence it increases \OM as well. This means that with always-one-bit-flip policy the algorithm behaves absolutely the same as on \OneMax (which also implies that its expected runtime is $(1 \pm o(1))n\ln(n)$).
We also find it interesting to show also algorithm's behavior when it chooses rates that are optimal for the optimization of \LeadingOnes as they were shown in~\cite{Doerr19domi,DoerrW18}. In our notation they are rates from the $\calSone$ state space, but they might be sub-optimal when we use \OM for offspring selection. This case is shown in Figure~\ref{fig:loom-s1-loopt}. These rates are more than one for all $\LO < \frac{n}{2}$, and it results in steeper trajectories. We conjecture that with such high rates it is much harder for the algorithm to optimize \OneMax without decreasing \LO value, and for this reason it is slower in the first half of optimization.

\textbf{Comparison of runtimes.} When we know all expected runtimes conditional on starting in each state, we can also compute the total expected runtime, which takes into account the initialization with a random bit string $x_0$. For this we can use the law of total expectation. To compute the probability that we start in state $s_{i, j}$ we first compute the probability to start with $\OM(x_0) = j$, which is $\binom{n}{j}2^{-n}$, since $\OM(x_0)$ follows a binomial distribution $\Bin(n, \frac{1}{2})$. Conditional on that, we note that there are $\binom{n}{j}$ ways to choose $j$ positions for one-bits (each of these ways has the same probability), and only $\binom{n - i - 1}{j - i}$ of them start with $1^i0$ (which is the same as having $\LO(x_0) = i$), hence the probability to start in state $s_{i, j}$ is
\begin{align*}
    \binom{n}{j}2^{-n} \cdot \frac{\binom{n - i - 1}{j - i}}{\binom{n}{j}} = \binom{n - i - 1}{j - i}2^{-n}.
\end{align*} 

\begin{figure}
    \begin{center}
        \begin{tikzpicture}
            \begin{axis}[width=0.95\linewidth, height=0.3\textheight,
                cycle list name=tikzcycle, grid=major,  xmode=log, log base x=2,
                legend style={at={(0.98,0.4)},anchor=south east},
                legend cell align={left}, ymax=2, ymin=0.5,
                xlabel={Problem size $n$}, ylabel={Expected runtime $/n\ln(n)$}]
                \addplot coordinates {(4,0.7213475204444817)(8,0.9429489852754729)(16,0.9691123046545619)(32,0.9710341715398176)(64,0.9739979802157617)(128,0.9769106238426685)(256,0.9794452633187157)(512,0.9815727333924914)(1024,0.9833450503559588)};
                \addlegendentry{Always one-bit flips};
                \addplot coordinates {(4,0.7889738504861519)(8,1.1004459642365665)(16,1.3198038019325313)(32,1.5478443294861148)(64,1.854223005957946)(128,2.279597910929133)(256,2.891750934952961)(512,3.788092492330035)};
                \addlegendentry{Optimal for \LO};
                \addplot coordinates {(4,0.6199080253819765)(8,0.8415295951187236)(16,0.9068397341167463)(32,0.9303317049752133)(64,0.9436745578753213)(128,0.9518070418038248)};
                \addlegendentry{Optimal for \loom};

            \end{axis}
        \end{tikzpicture}
    \end{center}
    \caption{The expected runtimes of the \rlsk on \loom with different methods of selecting the search radius. The runtimes are normalized by $n\ln(n)$. Note that the minimum $y$ value is not zero, it is made for a better visibility of the differences between algorithms.}
    \label{fig:loom-runtimes}
\end{figure}
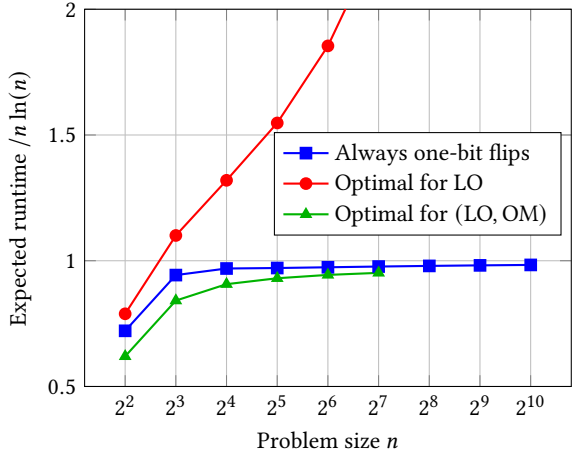

To compare the performance of the approaches we used, in Figure~\ref{fig:loom-runtimes} we show the plots of the total expected runtime of the \rlsk using the computed radii depending on problem size $n$. Those runtimes are normalized by $n\ln(n)$, since this is the asymptotical runtime when we always flip one bit, as we discussed before. This normalization helps us better see the difference between the performances of algorithms. We observe that the mutation rates that are optimal for \LeadingOnes are clearly not optimal for this function, and they yield runtime which is likely to be asymptotically larger than $n \ln(n)$. The runtime with the optimal policy which we computed is very similar to the runtime with only one-bit flips, however it is notably better. We assume that it is impossible to get a large advantage over the one-bit-flips policy, since the algorithm quickly gets into states, where it is \emph{optimal} to flip one-bits (with $\OM(x) > n/2$). However, the difference in the optimization speed at the very beginning of the process exists. It makes this benchmark an interesting and challenging testing ground for learning algorithms, since they need to learn the optimal rates having such a small reward for that.

\subsection{State Space of Bit Strings}
\label{sec:loom-sn}

In this subsection, we discuss the optimal policy for the state space $\calSn$, that is, when we can choose $k$ depending on the current bit string, but not only on its fitness value. Our goal is to find $K = (k_x)_{x \in \{0,1\}^n \setminus \{1^n\}}$ that minimizes the expected runtime. More precisely, for all $x \in \{0,1\}^n$ let $T_x(K)$ be the number of iterations that it takes the \rlsk to find the all-ones bit string when it starts in $x$ and chooses the search radii in each state according to $K$. Let also $p_s(x)$ be the probability to start in bit string $x$. Then the total expected runtime is
\begin{align*}
    E[T] = \sum_{x \in \{0, 1\}^n} p_s(x) E[T_x(K)],
\end{align*}
and we aim at finding $K$ that minimizes it.

The main difficulty in this case is in the possibility of transitions between bit strings that belong to the same state in $\calStwo$. This means that there are loops in the search space that complicate the computation of optimal $K$ (similar to what we will encounter in Section~\ref{sec:lo}, but much harder to solve due to the large size of the state space). Since our aim is to design benchmarks for which it is easy to compute the optimal $K$, we want to avoid such difficulties. Hence, we use the \rlsk with strict selection, that is, we assume that in line~\ref{line:comparison} of Algorithm~\ref{alg:rlsk} the inequality is strict. This implies that for any bit string $y$ that belongs to a state $s_{i, j} \in \calStwo$ the expected runtime $E[T_y(K)]$ depends only on the runtimes $E[T_z(K)]$ for bit strings $z$ from strictly larger states in $\calStwo$ than $s_{i, j}$, but not from $s_{i, j}$ itself. Hence, we can compute the expected runtimes for all states in $\calSn$ if we do it in order that corresponds to a descending order of states in $\calStwo$ (and it does not matter in which order we compute runtimes for bit strings inside the same $s_{i, j}$). Namely, similar to eq.~\eqref{eq:ET-loom-s2}, considering an arbitrary bit string $x$ from some state $s_{i, j} \in \calStwo$ and defining $\pleave$ as the probability to leave this bit string, we obtain
\begin{align*}
    E[T_x(K)] &= 1 + \sum_{(\ell, m): s_{\ell, m} > s_{i, j}} \sum_{y \in s_{\ell, m}} p_{x \to y}(k_x) E[T_y(K)] \\
    &+ (1 - \pleave) E[T_x(K)]; \\
    E[T_x(K)] &= \frac{1 + \sum_{(\ell, m): s_{\ell, m} > s_{i, j}} \sum_{y \in s_{\ell, m}} p_{x \to y}(k_x) E[T_y(K)]}{\pleave}.
\end{align*}

\begin{figure}
    \includegraphics[width=0.95\linewidth]{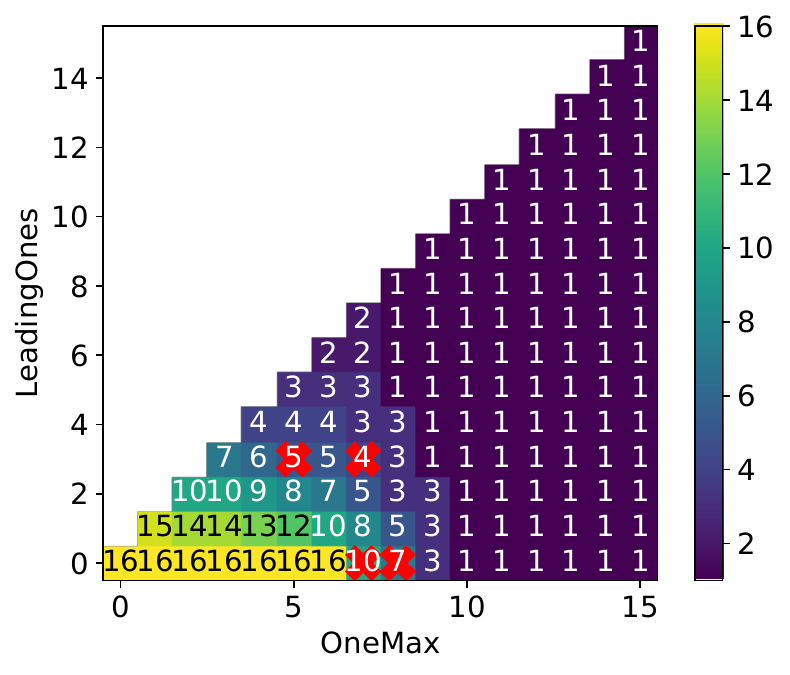}
    \caption{Optimal rates of the \rlsk on \loom for $n = 16$. The states in which some bit strings have a different optimal rate in $\calSn$ are marked with red crosses.}
    \label{fig:loom-sn}
\end{figure}

Using this equation, we can compute the optimal policy for each bit string $x$ in the search space by trying all values of $k_x$ that can give progress for this bit string and finding the one which minimizes $T_x(K)$. However, since the search space grows exponentially with the growth of the problem size $n$, the maximum $n$ for which we could compute the optimal rates is $n = 16$.  For this problem size $n = 16$ we have found that for most bit strings their optimal search radius was the same as in $\calStwo$ (but with a standard non-strict selection used in the \rlsk). Figure~\ref{fig:loom-sn} shows the optimal rates for $\calStwo$ for $n = 16$ and marks the cells, where there are bit strings with a different optimal radius. There are only four states for which there is a difference in optimal rates, and all those states are in areas where the optimal rate in $\calStwo$ changes most quickly (that is reflected with the strongest gradient in the heatmap).

The results of this section indicate that we do not benefit much from having information about the precise bit string, and the optimal rates for almost all bit strings are the same as in the state space $\calStwo$. Nevertheless, due to the large state space this is a very interesting setting for testing learning agents. Moreover, for such a large state space the set of close-to-optimal rates that we can use as a baseline can be computed as in Section~\ref{sec:loom-s2}, which can be done relatively fast even for extremely large state space $\calSn$.  

\section{Optimization of \LeadingOnes}
\label{sec:lo}

\begin{figure*}
    \includegraphics[width=0.33\linewidth]{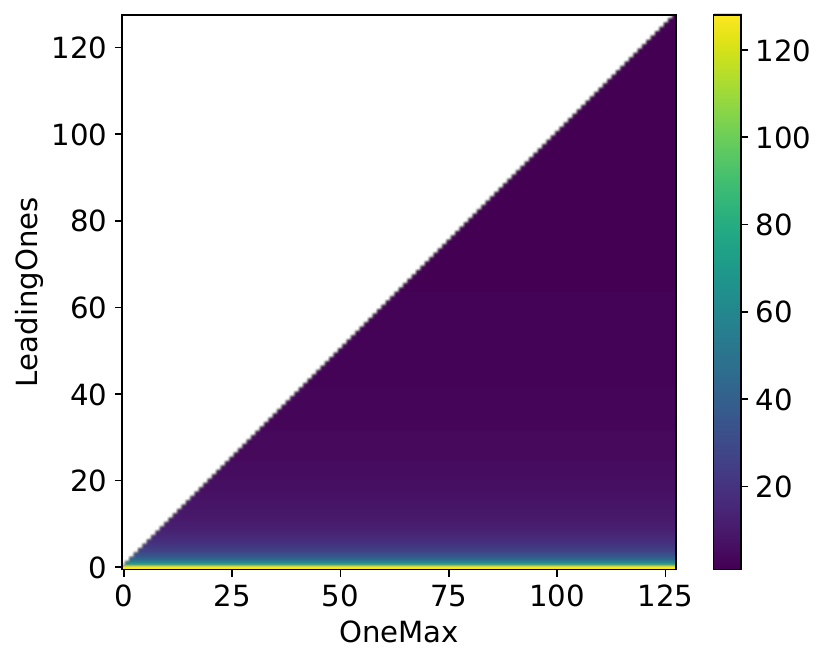}
    \includegraphics[width=0.33\linewidth]{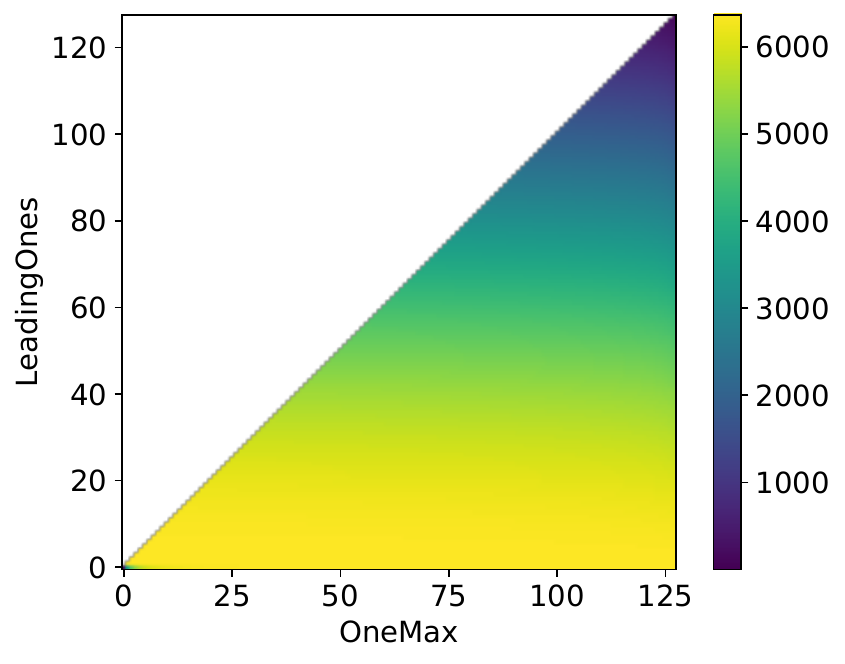}
    \includegraphics[width=0.33\linewidth]{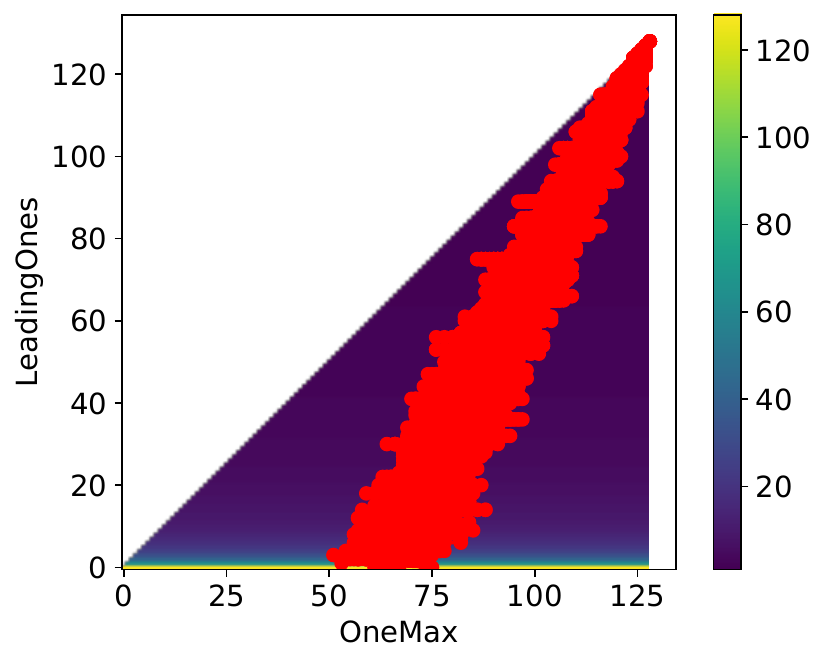}
    \caption{Heatmaps for the \rlsk on \LeadingOnes with optimal \LO-based search radii: the radii themselves (left), expected runtimes (middle) and trajectories of 10 typical runs over the optimal rates heatmap (right). The problem size is $n = 128$.}
    \label{fig:lo-s1}
\end{figure*}

As it has been mentioned, it is well-known from~\cite{Doerr19domi,DoerrW18} that the optimal fitness-dependent mutation rate for \LeadingOnes is $k = \lfloor\frac{n}{\LO(x)}\rfloor$, where $x$ is the current individual, hence we use this setting as a baseline. The heatmaps describing this case are shown in Figure~\ref{fig:lo-s1}.

\begin{figure*}[ht]
    \includegraphics[width=0.33\linewidth]{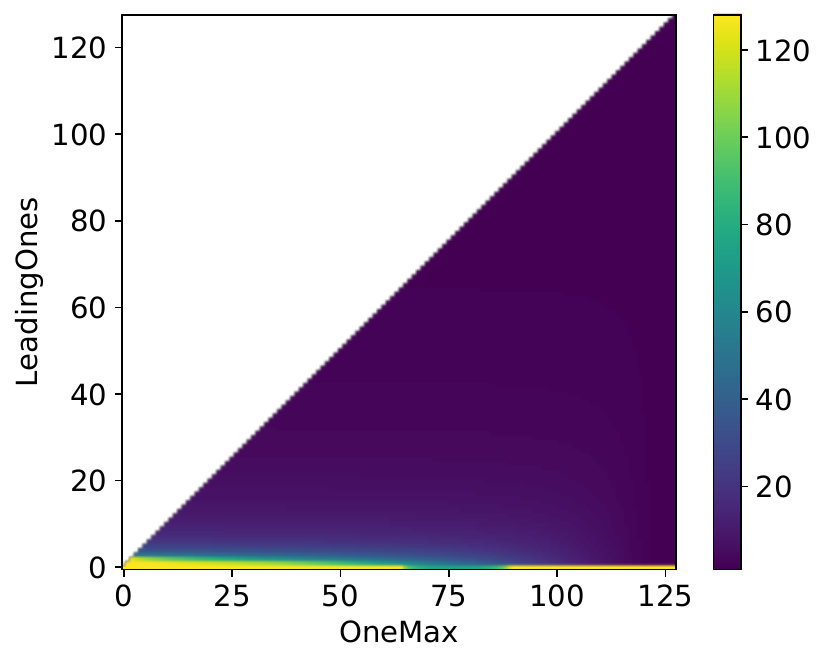}
    \includegraphics[width=0.33\linewidth]{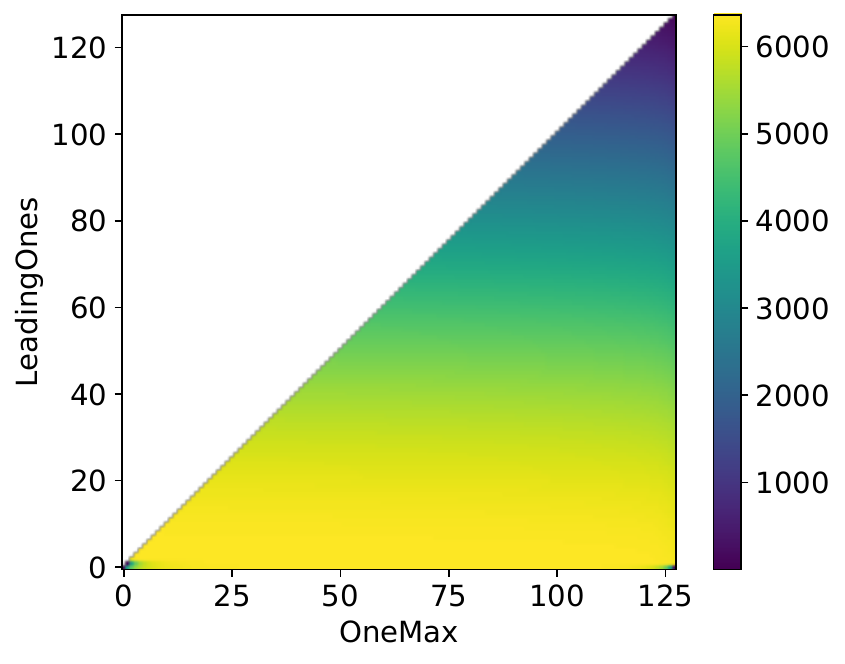}
    \includegraphics[width=0.33\linewidth]{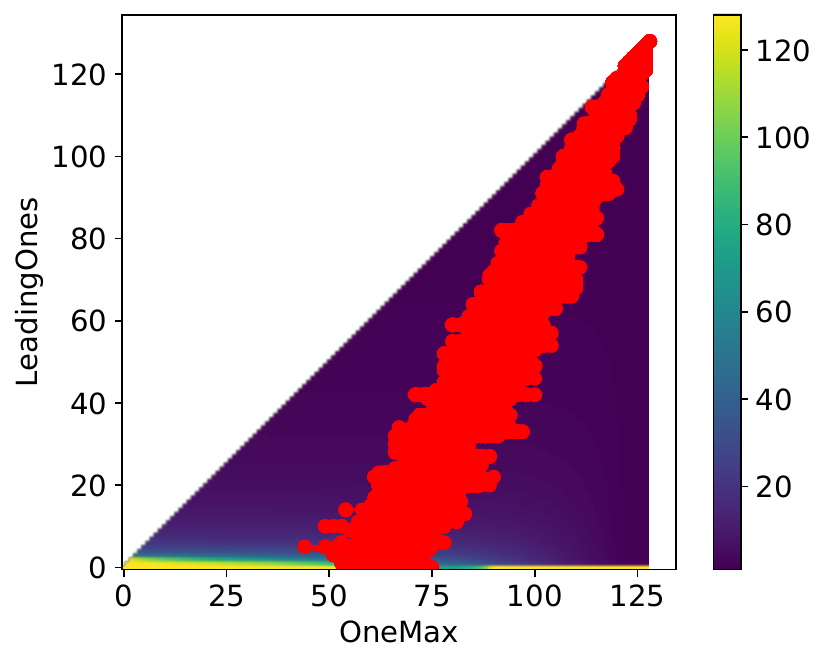}
    \caption{Heatmaps for the \rlsk on \LeadingOnes with approximated optimal policy based on the $\calStwo$ state space: optimal search radii (left), expected runtimes (middle) and trajectories of 10 typical runs over the optimal rates heatmap (right). The problem size is $n = 128$.}
    \label{fig:lo-s2}
\end{figure*}

\begin{figure*}[ht]
    \includegraphics[width=0.33\linewidth]{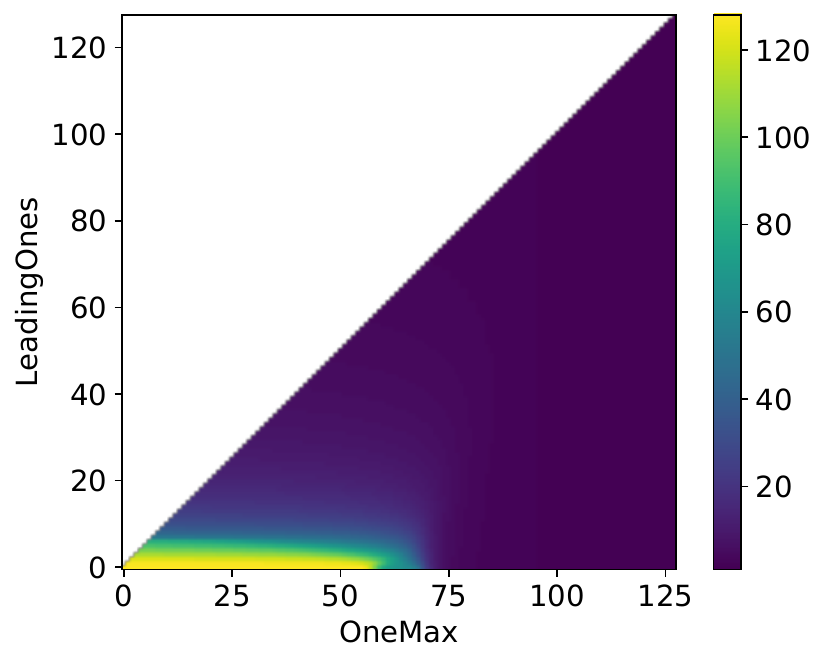}
    \includegraphics[width=0.33\linewidth]{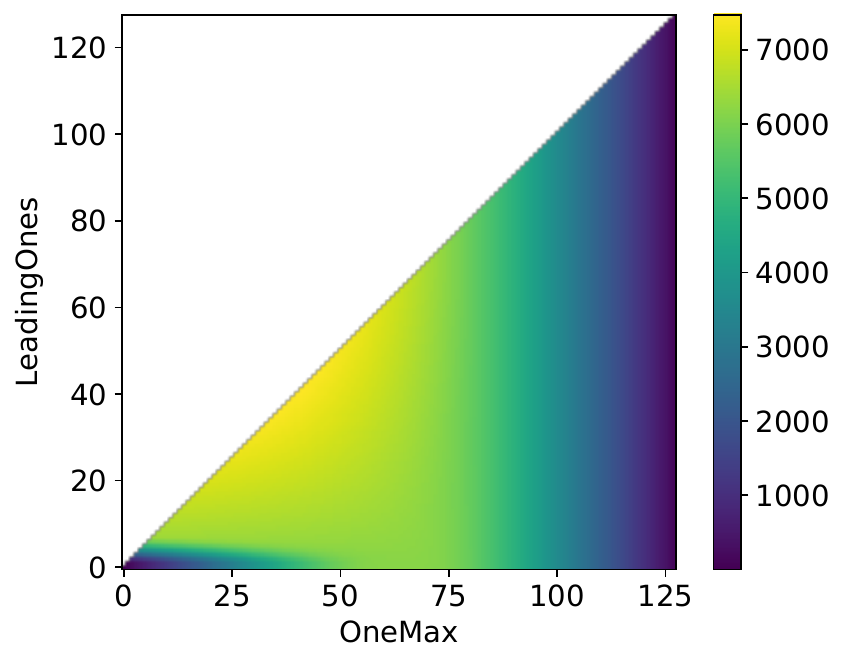}
    \includegraphics[width=0.33\linewidth]{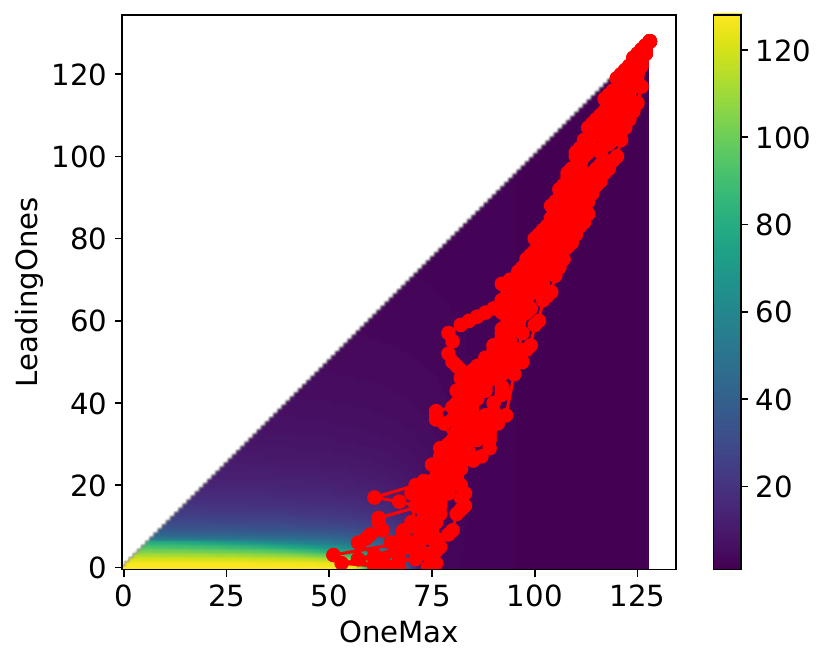}
    \caption{Heatmaps for the \rlsk with strict selection on \LeadingOnes with optimal policy based on the $\calStwo$ state space: optimal search radii (left), expected runtimes (middle) and trajectories of 10 typical runs over the optimal rates heatmap (right). The problem size is $n = 128$.}
    \label{fig:lo-strict}
\end{figure*}

In this section we introduce two benchmarks based on \rlsk optimizing \LeadingOnes (without additional \OM information for offspring selection) and using $\calStwo$ state space (that is, we can choose $k$ based on the current \LO and \OM values). The difference between the two benchmarks is in the selection used in the \rlsk: in one case it is standard selection which accepts offspring that are not worse than the parent, and in the second case it is strict selection that accepts only strictly improving offspring. The main question we aim to answer in this section is if introducing an additional information (in our case, the current \OneMax value) can help \rlsk to choose better radii in some states and if such choices can affect the optimization time enough for the learning methods to be able to find the optimal rates. 

\subsection{Standard Selection}
\label{sec:lo-standard}

We start with computing the optimal radius for each state in $\calStwo$ state space for the standard non-strict selection. This is not a trivial task. The algorithm's selection is now not based on the value of \OneMax, hence it can make loops in the state space inside one \LeadingOnes level. This implies that there is no order of states so that the algorithm could only go from smaller states to larger states, which does not allow us straightforwardly compute the best rates as we did it in Section~\ref{sec:loom}. Instead, we use the following approach.

Similar to Section~\ref{sec:loom-s2}, let $T_{i, j}(K)$ be the expected runtime when we start from state $s_{i,j}$ and use policy $K$. Assume that we have computed optimal rates and runtimes for all \LeadingOnes levels above some arbitrary level $i \in [0..n-1]$. Then let $K$ be some fixed policy with optimal rates $k_{\ell,m}$ for all $\ell > i$ and some arbitrary fixed rates for all other states. We can compute the expected runtimes for all states with $\LO = i$ by solving a system of $n - i$ linear equations, where each equation corresponds to one of these states. The equation corresponding to $s_{i, j}$ is as follows.

\begin{align}\label{eq:ET-lo-s2}
    \begin{split}
        E[T_{i, j}(K)] &= 1 + \sum_{\ell = i}^{n} \sum_{j=\ell}^{n} p_{i, j}^{\ell, m}(k_{i, j}) E[T_{\ell, m}(K)] \\
        &+ (1 - \pleave) E[T_{i, j}(K)]; \\
        \pleave E[T_{i, j}(K)] &- \sum_{j=i}^{n} p_{i, j}^{i, m}(k_{i, j}) E[T_{i, m}(K)] \\
        &= 1 + \sum_{\ell = i + 1}^{n} \sum_{j=\ell}^{n} p_{i, j}^{\ell, m}(k_{i, j}) E[T_{\ell, m}(K)].
    \end{split}
\end{align}
Left side of the last equations contains the expected runtimes for the states with \LO value $i$ (that we want to compute now), while in the right side we gather the terms from higher \LO levels, which by our assumption is already computed and is optimal. Solving this system gives us expected runtimes for states $s_{i, i},\dots, s_{i, n - 1}$ and for policy $K$, however to find the best policy we need to check all possible combinations of values of $k_{i, i}, \dots, k_{i, n - 1}$, which is extremely time-consuming. Even if we do not check $k > n - i$ (since this number of bits flipped reduces \LO value with probability one), the number of different combinations is $(n - i)^{(n - i)}$. It is also not clear if the optimal policy exists in this case: it might happen that one policy minimizes runtime in one state, while another policy minimizes it in another. 

To tackle this problem we developed the following heuristic method of computing the optimal rates. 
We first set all $k_{i,j} = 1$ (for all $j \in [i..n - 1]$) and find the first approximation of the optimal runtimes by solving system of equations as in eq.~\eqref{eq:ET-lo-s2}. Let the resulting \emph{expected} runtimes be $T_{i, i}^{(0)}, \dots, T_{i, n - 1}^{(0)}$. Then for each $j \in [i..n-1]$ in eq.~\eqref{eq:ET-lo-s2} we replace all $E[T_{i, m}(K)]$ ($m \ne j$) with corresponding $T_{i, m}^{(0)}$ and obtain
\begin{align*}
    E[T_{i, j}(K)] = \frac{1}{\pleave} \cdot \bigg(1 &+ \sum_{j=i}^{n} p_{i, j}^{i, m}(k_{i, j}) T_{i, m}^{(0)}(K) \\
    &+ \sum_{\ell = i + 1}^{n} \sum_{j=\ell}^{n} p_{i, j}^{\ell, m}(k_{i, j}) E[T_{\ell, m}(K)]\bigg).
\end{align*}
We now can find the value of $k_{i, j}$ that minimizes this expression of $E[T_{i, j}(K)]$ by trying all values from the portfolio. After finding all $k_{i, j}$ we re-solve the system and obtain a new approximation of optimal runtimes $T_{i, i}^{(1)}, \dots, T_{i, n - 1}^{(1)}$. We repeat this process until all values of $k_{i, j}$ are not changed after an iteration of this process (that is, the process converged to some vector of $k$-s). In each iteration we need to check $(n - i)$ values for each of $(n - i)$ values $k_{i, j}$, hence we check $(n - i)^2$ different vectors combinations if $k_{i, j}$. This turns to be much faster than the brute force search, and it takes at most 5 iterations for the process to converge for problem sizes that we tried (up to $n = 256$).

\begin{figure}
    \begin{center}
        \begin{tikzpicture}
            \begin{axis}[width=0.95\linewidth, height=0.2\textheight,
                cycle list name=tikzcycle, grid=major,  xmode=log, log base x=2, legend pos=south east,
                legend cell align={left},
                xlabel={Problem size $n$}, ylabel={Expected runtime $/ n^2$}]
                
                \addplot coordinates {(4,0.2994791666666667)(8,0.3750065902693765)(16,0.38596651471823235)(32,0.3875143628368094)(64,0.38803236473614383)(128,0.3882592614981722)};
                \addlegendentry{$\calSone$-optimal};
                \addplot coordinates {(8,0.3399544324444243)(16,0.37944896466522865)(32,0.38655490379539204)(64,0.38790877308544774)(128,0.3882423979647848)(256,0.3883566447000756)};
                \addlegendentry{$\calStwo$-optimal};
            \end{axis}
        \end{tikzpicture}
    \end{center}
    \caption{The expected runtimes of the \rlsk on \LeadingOnes with different methods of selecting the search radius. The runtimes are normalized by $n^2$. Note that the minimum $y$ value is not zero, it is made for a better visibility of the differences between algorithms.}
    \label{fig:lo-runtimes}
\end{figure}
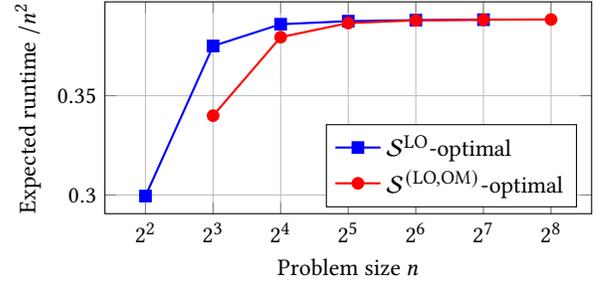

The heatmaps showing the search radii that are computed with this method, expected runtimes they yield and trajectories of typical runs are shown in Figure~\ref{fig:lo-s2}. The picture is very similar to Figure~\ref{fig:lo-s1}. In particular, in this figure we see that the trajectory of runs is well concentrated around states with $\OM = \frac{n + \LO}{2}$, that is, where we have approximately half ones and half zeros in the suffix. In those states the optimal rates are almost always the same as the optimal rates based only on \LeadingOnes value from~\cite{Doerr19domi,DoerrW18}. This leads to the minimal difference from the performance gain from using \OM value to choose $k$. This is illustrated in Figure~\ref{fig:lo-runtimes}, where we compare the runtimes with optimal rates based only on \LO value ($\calSone$ state space) and based on \LO and \OM values ($\calStwo$ state space).

\subsection{Strict Selection}
\label{sec:lo-strict}

To overcome the computation complexity of the optimal $\calStwo$ policy for the \LO objective, we also consider a setting with strict selection embedded into \rlsk. In contrast with the previous subsection, it creates an order of states: the algorithm can transition from state $s_{i,j}$ to state $s_{\ell, m}$ only if $\ell > i$ (independently of \OM values $j$ and $m$). This allows us to use a method of computing the expected runtimes similar to the one from Section~\ref{sec:loom-s2}. We now have the following equation for the expectation of $T_{i, j}(K)$ which is similar to eq.~\eqref{eq:ET-loom-s2}. 
\begin{align}\label{eq:ET-lo-strict}
    \begin{split}
        E[T_{i, j}(K)] &= 1 + \sum_{\ell = i + 1}^{n - 1} \sum_{m = \ell}^{n - 1} p_{i, j}^{\ell, m}(k_{i, j}) E[T_{\ell, m}(K)] \\
        &+ (1 - \pleave) E[T_{i, j}(K)]; \\
        E[T_{i, j}(K)] &= \frac{1 + \sum_{\ell = i + 1}^{n - 1} \sum_{m = \ell}^{n - 1} p_{i, j}^{\ell, m}(k_{i, j}) E[T_{\ell, m}(K)]}{\pleave}.
    \end{split}
\end{align}
Note that there is no term for state $s_{n, n}$ in this equation, since $E[T_{0, 0}(K)] = 0$, but the transition probability $p_{i, j}^{n, n}(k_{i, j})$ is included into $\pleave$.

\begin{figure}
    \begin{center}
        \begin{tikzpicture}
            \begin{axis}[width=0.95\linewidth, height=0.2\textheight,
                cycle list name=tikzcycle2, grid=major,  xmode=log, log base x=2, legend pos=south east,
                legend cell align={left},
                xlabel={Problem size $n$}, ylabel={Expected runtime $/ n^2$}]
                
                \addplot coordinates {(8,0.3399544324444243)(16,0.37944896466522865)(32,0.38655490379539204)(64,0.38790877308544774)(128,0.3882423979647848)(256,0.3883566447000756)};
                \addlegendentry{Standard selection};
                \addplot coordinates {(8,0.2990749395461309)(16,0.3316488202904247)(32,0.3502670907258321)(64,0.3642475244316819)(128,0.37366356566874476)(256,0.379750730734364)};
                \addlegendentry{Strict selection};
            \end{axis}
        \end{tikzpicture}
    \end{center}
    \caption{The expected runtimes of the \rlsk on \LeadingOnes with optimal search radii for strict and non-strict offspring selection. The runtimes are normalized by $n^2$. Note that the minimum $y$ value is not zero, it is made for a better visibility of the differences between algorithms.}
    \label{fig:lo-runtimes-strict}
\end{figure}
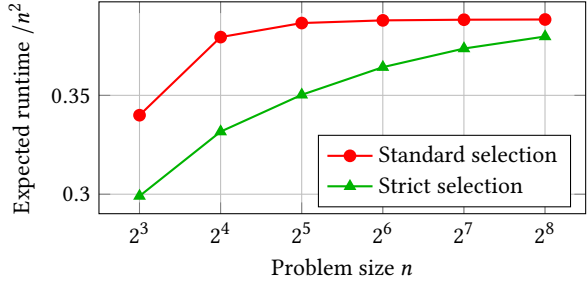

With eq.~\eqref{eq:ET-lo-strict} we can compute the optimal rates and the expected runtimes for each state. We illustrate them with heatmaps in Figure~\ref{fig:lo-strict} for problem size $n = 128$. In this figure one can see a significant difference in the optimal rates, runtimes and behavior of the \rlsk compared to the standard selection (Figure~\ref{fig:lo-s2}). The first interesting trend that we observed is that the optimal $k$ is $k_{i,j} = 1$ for all states with \OM value $j \ge 96$, and it is larger than one for all other states. This is also reflected in expected runtimes: we observe a gradient along the $X$-axis in this one-bit-flip area, but no such gradient along the $Y$-axis, as was seen in Figure~\ref{fig:lo-s2}. This is because once the \rlsk is in the zone of one-bit flips, it can only replace the parent $x$ by flipping the first zero-bit in it. Hence, each replacement of the parent takes $n$ iterations in expectation and it always increases \OM value by one. Consequently, the expected time until the algorithm finds the optimum starting in a state $s_{i, j}$ from this one-bit-flip zone is exactly $jn$, that is, it is the same for all states with the same \OM value.

Another notable difference with standard offspring selection is that the total expected runtime with strict selection is notably smaller, which is illustrated in Figure~\ref{fig:lo-runtimes-strict}. While it looks like the relative difference decreases as the problem size grows, it seems to be large enough to give a good signal to learning agents. 

It is also interesting to mention that the worst-case initial states are very similar to the worst-case states for optimizing \loom: they are the states with $\OM \approx \LO \le \frac{n}{3}$, with the worst state being $s_{41, 41}$. The expected runtime for that state is $7469.51$ iterations, which is larger than for the worst-case starting state for standard selection, which is $6362.68$ (for state $s_{3, 3}$). This is surprising in the light of the total expected runtime being better for the strict selection.  

We conjecture that the reason for the advantage of the strict selection can be explained as follows. When the \rlsk uses \emph{standard selection}, it can change the \OM value without improving \LO value. It leads to the drift of \OM value towards $\frac{n + \LO(x)}{2}$, that is, towards the equal number of ones and zeros in the tail. This drift is strong enough for the algorithm to reach this area before creating an individual with better \LO value, which means that essentially we always use the optimal rates for those states for improvements, and those rates are very similar to the optimal rates based only on \LO value. With \emph{strict selection}, if at some point the \OM value fluctuates from $\frac{n + \LO(x)}{2}$, it stays there until we improve \LO value. In the very early stages of optimization such fluctuations towards smaller number of one-bits lead us to the area where we can get large progress with many-bit flips (the bottom left area in the heatmaps in Figure~\ref{fig:lo-runtimes-strict}), and in slightly later stages of optimization it can lead us to the one-bit-flip area on the right, where the algorithm does not lose the accumulated one-bits in the suffix of $x$.

These observations create a very interesting landscape of optimal rates and of the behavior of the algorithm, which is a promising testing ground for learning mechanisms.

\section{Limited portfolio}
\label{sec:portfolio}

Lastly, we calculated the expected runtime for the \rlsk with standard selection optimizing \loom when only a limited portfolio of search radii is available. Taking inspiration from the paper by~\citet{BiedenkappDKHD22} where limited portfolios are used to train the RL-agent, we computed the exact runtime for the portfolios consisting of powers of two, i.e., $\{2^i \mid 2^i \leq n\}$, initial segment with 3 elements, i.e., $[1..3]$, and evenly spread portfolio with 3 elements, i.e., $\{1, \lfloor \frac{n}{3} \rfloor, \lfloor \frac{2n}{3} \rfloor \}$.

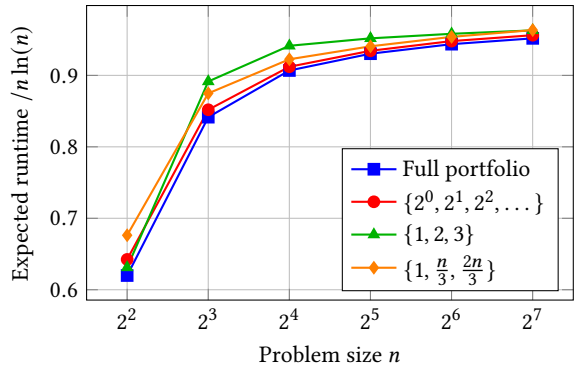
\begin{figure}
    \begin{center}
        \begin{tikzpicture}
            \begin{axis}[width=0.95\linewidth, height=0.25\textheight,
                cycle list name=tikzcycle, grid=major,  xmode=log, log base x=2, legend pos=south east,
                legend cell align={left},
                xlabel={Problem size $n$}, ylabel={Expected runtime $/n\ln(n)$}]

                \addplot coordinates {(4,0.6199080253819765)(8,0.8415295951187236)(16,0.9068397341167463)(32,0.9303317049752133)(64,0.9436745578753213)(128,0.9518070418038248)};
                \addlegendentry{Full portfolio};
                \addplot coordinates {(4,0.6424501353958665)(8,0.8518724638923594)(16,0.912055165165285)(32,0.9343968939172913)(64,0.9479737382389717)(128,0.9562184961461839)};
                \addlegendentry{$\{2^0, 2^1, 2^2, \dots\}$};
                \addplot coordinates {(4,0.6311790803889216)(8,0.8914739518626114)(16,0.9414098465768493)(32,0.9518152887801266)(64,0.9581812689781305)(128,0.9629823554566624)};
                \addlegendentry{$\{1, 2, 3\}$};
                \addplot coordinates {(4,0.6762633004167016)(8,0.8748397954701824)(16,0.9225242918350762)(32,0.9406636769285354)(64,0.9538927210276328)(128,0.9633879669349147)};
                \addlegendentry{$\{1, \frac{n}{3}, \frac{2n}{3}\}$};
            \end{axis}
        \end{tikzpicture}
    \end{center}
    \caption{The expected runtimes of the \rlsk on \loom with different portfolios. The runtimes are normalized by $n\ln(n)$. Note that the minimum $y$ value is not zero, it is made for a better visibility of the differences between algorithms.}
    \label{fig:portfolios-runtimes}
\end{figure}

The best possible runtimes of the \rlsk with these portfolios are shown in Figure~\ref{fig:portfolios-runtimes}. We computed them in the similar way as in Section~\ref{sec:loom-s2}. It is interesting to note that the portfolio choice does not seem to strongly affect the performances, and the difference between normalized runtimes decreases with the growth of problem size, however we can see that portfolio $\{1, 2, 3\}$ has the worst performance. It is likely because it does not have large radii that are beneficial in the early steps of the optimization.

\section{Summary of the Results}
\label{sec:summary}

\begin{table*}
    \centering
    \begin{tabular}{cccm{8cm}c}
    \toprule
        Fitness function & Offspring selection & State space & Description & Max $n$ computed  \\\midrule
        \multirow{2}{*}{\loom}  &
        Non-strict &
        $\calStwo$ &
        We can compute the optimal policy precisely in polynomial time. The optimal policy is shown in Figure~\ref{fig:loom-s2} and it gives a better runtime that one-bit-flip policy. & 
        $n = 256$ \\\cmidrule(lr){2-5}
        
        & Strict &
        $\calSn$ &
        The state is exponentially big, we cannot effectively compute the policy. For $n=16$ almost always the optimal search radius is the same as for state space $\calStwo$. &
        $n = 16$ \\\midrule

        \multirow{2}{*}{\LeadingOnes}  &
        Non-strict &
        $\calStwo$ &
        We cannot effectively compute the optimal policy, but can approximate it by relaxing the optimal values of $k$ iteratively for each \LO value. The resulting rates are shown in Figure~\ref{fig:lo-s2}. We get a slightly better runtime than with optimal search radii that depend only on \LO.  & 
        $n = 256$ \\\cmidrule(lr){2-5}
        
        & Strict &
        $\calStwo$ &
        We can compute the optimal policy in polynomial time. It is shown in Figure~\ref{fig:lo-strict}. The resulting runtime has a better-pronounced advantage over the \LO-based optimal policy. &
        $n = 16$ \\        
    \bottomrule
    \end{tabular}
    \caption{The summary of the proposed benchmarks. The last column indicates for which maximum $n$ we computed the optimal policy. Note that in the paper most heatmaps are for $n = 128$, which is not the maximum one, but it better illustrates some details.}
    \label{tbl:results}
\end{table*}

In this work we proposed the four benchmarks that are described in Table~\ref{tbl:results}. For the results based on \LeadingOnes function (with no clues for offspring selection, see Section~\ref{sec:lo}) we observed that adding additional information can improve the runtime by allowing the algorithm to make better choices. These choices, however, have a limited effect on the performance, since typically the algorithm's trajectory goes through the states with $\OM \approx \frac{n + \LO}{2}$. This effect of the extended state space can be amplified by using strict offspring selection in \rlsk, since it allows the algorithm to stay longer in marginal states where \OM value is further away from $\frac{n + \LO}{2}$ and where a non-typical search radius can be beneficial.

Studying \loom, which uses additional hints from \OneMax in offspring selection, we first computed optimal rates as a baseline. These rates resulted into $O(n\log(n))$ runtime of the algorithm, slightly improving the policy of always flipping one bit. What makes this setting interesting for testing DAC methods is the huge decline of performance which we observed when chose non-optimal policy. In this case, however, further extension of the state space did not give any effect: the best policy of choosing search radius for each bit string was almost the same as the optimal policy based on the fitness. The positive side of this is that we can use the optimal policy for $\calStwo$, which is easy to compute, as a reference for testing learning methods that look for an optimal policy in the much larger state space $\calSn$.

We are convinced that examples like ours can be very useful in the context of dynamic algorithm configuration~\cite{Adriansen2022DACjournal}, where they can be exploited as benchmarks with known ground truth~\cite{BiedenkappDKHD22,DeyaoGECCO23}. In this spirit, we plan to investigate the impact of different state space choices for other settings, in particular for other problems, other algorithms, and other performance criteria. By replacing actual algorithm runs by simulated ones, it is also possible to scale the stochasticity of the reward that is assigned to a chosen action. Such a treatment and gradual decrease of the randomness would deliver relevant insights into the brittleness of dynamic algorithm configuration approaches. 

\begin{acks}
The project is financially supported by Alliance Sorbonne Université (project number EMERGENCE 2023 RL4DAC), by the European Union (ERC CoG ``dynaBBO'', grant no.~101125586) and by ANR project ANR-23-CE23-0035 Opt4DAC. This work used the supercomputer at MeSU Platform (\href{https://sacado.sorbonne-universite.fr/plateforme-mesu}{https://sacado.sorbonne-universite.fr/plateforme-mesu}). Our work benefited from discussions with COST Action CA22137 ``Randomized Optimization Algorithms Research Network'' (ROAR-NET), supported by the European Cooperation in Science and Technology.
\end{acks}

\bibliographystyle{ACM-Reference-Format}
\bibliography{references}

\end{document}